# Learning from Noisy Labels with Distillation

Yuncheng Li    Jianchao Yang    Yale Song    Liangliang Cao    Jiebo Luo    Li-Jia Li

## Abstract

*The ability of learning from noisy labels is very useful in many visual recognition tasks, as a vast amount of data with noisy labels are relatively easy to obtain. Traditionally, label noise has been treated as statistical outliers, and techniques such as importance re-weighting and bootstrapping have been proposed to alleviate the problem. According to our observation, the real-world noisy labels exhibit multi-mode characteristics as the true labels, rather than behaving like independent random outliers. In this work, we propose a unified distillation framework to use "side" information, including a small clean dataset and label relations in knowledge graph, to "hedge the risk" of learning from noisy labels. Unlike the traditional approaches evaluated based on simulated label noises, we propose a suite of new benchmark datasets, in Sports, Species and Artifacts domains, to evaluate the task of learning from noisy labels in the practical setting. The empirical study demonstrates the effectiveness of our proposed method in all the domains.*

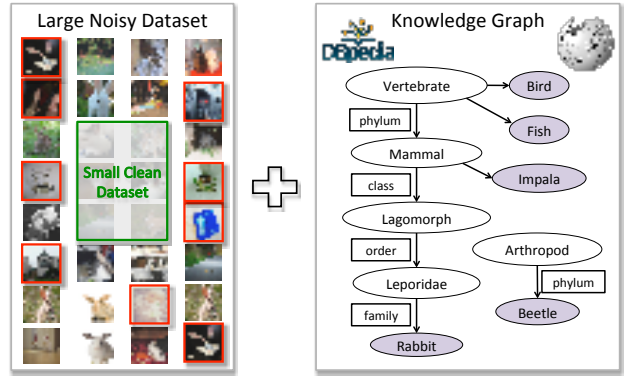

Figure 1: Overview of the proposed system to learn from noisy labels by leveraging a knowledge graph. The left panel shows the large scale noisy dataset, out of which we collect a small set of images with clean labels to guide the learning process. On the right panel, we demonstrate the knowledge graph on the species domain constructed from DBpedia-Wikipedia.

## 1. Introduction

With the recent advancements in deep convolutional neural networks (CNN) [11], learning from a clean large-scale dataset, e.g., ImageNet [17], has been very successful in various visual recognition tasks. However, collecting such datasets is time consuming and expensive. Recent efforts, therefore, have been focused on building and learning from an Internet-scale dataset with noisy labels such as YFCC100M [20] and YouTube8M [1]. These datasets have the potential of leveraging a seemingly infinite amount of images and videos on the Internet. But labels in those datasets are noisy in terms of visual correlation and hence challenging for the learning process.

Previous approaches tried to circumvent the problem of learning from noisy samples by treating them as statistical outliers and discarding them using some variants of outlier detection methods [16, 12, 18]. However, in practice, it is typical that noisy samples are not statistical outliers but rather some form of significant mass. Existing approaches have shown to produce inferior results on these cases. For ex-ample, images collected by searching polysemy words, such as *apple*, will show a multi-modal distribution of visual concepts, in which case statistical outlier detection techniques will fail to figure out which concept to be associated with. Another example, images labeled with *basketball* on Flickr may contain a significant amount of group shots, selfies, and photos taken before or after the game – these are less visually relevant to the event itself but regardless forming a significant mass; statistically, they are not outliers.

Recently, Hinton *et al*. [9] introduced the concept of "distillation" to transfer the knowledge learned from one model (expert or teacher model) to another (a lightweight student model), by treating the prediction results produced from the first model (usually more expensive to train) as the "soft target" labels for training the second light model (usually trained in a more constrained setting). Inspired by this, we propose a new technique that uses a similar distillation process to learn from noisy datasets. In our scenario, we assume that we have a small clean dataset and a large noisy dataset. The small clean dataset can either be an existing public dataset or labeled from part of the noisy data. Our goal is to use the large amount of noisy data to augment the small clean dataset to learn a better visual representation and classifier. Concretely, we distill the knowledge learned the small clean dataset to facilitate learning a better model from the entire noisy dataset. This is different from Hinton *et al.* [9], where distillation is used to transfer knowledge from a better model (e.g., an ensemble model) to guide learning a light but typically inferior model. Furthermore, we propose to integrate a knowledge graph to guide the distillation process, where rich relational information among labels are explicitly encoded in the learning process. This helps the algorithm to disambiguate noisy labels by, e.g., knowing that apple can either be a fruit category or a company name.

[1]https://research.google.com/youtube8m/

To evaluate our technique, we collect a suite of new datasets on three topics: *sports*, *species*, and *artifacts*. Our dataset contains a total of 480K images from 780 class categories and exhibit the real-world labeling noise we mentioned above. We build a textual knowledge graph on top of these three topics based on Wikipedia, where labels are related by their definitions. We show that, our proposed distillation process, as well as leveraging the knowledge graph to guide the distillation process, can achieve the best results on our datasets compared with competing methods.

In summary, we make the following contributions:
- We propose a novel algorithm based on a distillation process to learn from noisy data, with a theoretical analysis under some conditions.
- We leverage a knowledge graph to guide the distillation process to further "hedge the risk" of learning from noisy labels.
- We collect several new benchmark datasets with real-world labeling noises. We extensively compare with different baselines and show that our proposed algorithm achieves the best results [2].

## 2. Related Work

In this section, we review related works on learning with noisy labels and network distillation.

Learning with noisy labels has been an important research topic since the beginning of machine learning. Recently, Reed *et al.* proposed a bootstrap technique to modify the labels on-the-fly, in order to alleviate the potential damage caused by the noisy labels [16]. Liu *et al.* proposed an importance re-weighting method to deal with random classification noise [12], where the level of label noise is estimated from a pretrained classifier. This approaches extends the idea of unbiased loss function [14] in the traditional importance re-weighting framework. Sukhbaatar *et al.* proposed a noise layer on top of softmax to "absorb" label noise [18]. Szegedy, *et al.* proposed a simple yet effective way to avoid over-trusting the noisy labels [19], by uniformly redistributing the energy of the noisy labels. Interestingly, Krause *et al.*

[2]The code and dataset will become publicly available after the paper is accepted.

discovered that, in the case of fine grained classification, the label noise does not hurt the performance, because the noisy examples are not shown in the evaluation [10].

Learning with noisy labels is an important technique with many applications [8] [4]. Especially a number of researchers studied the problem of learning from text based image search results. For example, Divvala *et al.* introduced a fully automatic system to learn visual concepts and their variations using image search results from the Internet [6]. Chen *et al.* proposed a two-step approach to learn *ConvNet* by image search on the Internet [3].

To handle the noisy labels, a number of researchers consider the classification with side information strategy to handle data noise and accelerate optimization. For example, Wu *et al.* proposed a framework with mixed graph to handle missing labels in the task of multi-label classification [22]. Bergamo *et al.* also proposed to exploit small manually labeled dataset to learn with text based image search in the framework of domain adaptation [1]. Frome *et al.* proposed to use the word2vec distance of the labels to scale the *ConvNet* learning process to a larger vocabulary size. These methods are thought-provoking, however, differ from our approach in the way of leveraging clean and noisy labels.

Sukhbaatar and Fergus [18] proposed a novel layer to handle the noisy labels in the context of neural network. After the network is trained with the baseline method, an extra linear layer is added on top of the network to "absorb" label switch noise. This new layer estimates the label switching probability with a linear function, which works well with simulated noise that conforms to the label switch assumption. However, it is not clear how well this approach works in real world scenarios. We will compare this method with ours in the experiments.

Our approach is motivated by the recent works of network distillation [9, 13]. Hinton *et al.* developed the idea of distillation [9] to learn a student model with simpler network structure to replace the teacher model with a cumbersome ensemble of models. Similarly, Bulò *et al.* used distillation to extract an optimal predictor from a model trained with dropout, which outperforms the standard scaling based dropout [2]. Lopez-Paz *et al.* unified distillation and privileged information into one framework [13]. A key insight from these works is that the soft distillation scores are better than hard labels when guiding the learning of student networks. However, as discussed before, our approach uses a different setting as traditional distillation approaches. Table 1 summarizes these key differences. In addition, traditional distillation approaches assume the teacher network has better performance with better empirical risk bounds [13]. In this paper, we show that the student network can do better than the teacher. Given a teacher network is trained from clean dataset, this paper proposes to leverage a bigger dataset with noisy labels with the outputs of teacher network, which leads to a student model that consistently outperforms the teacher.

## 3. Our method

In this section, we formulate the problem of learning from noisy labels based on distillation, and explain how we can further improve the learning process by using a knowledge graph as a guide to the distillation process.

### 3.1. Problem Formulation

Consider an $L$ way multi-class classification dataset,

$$\mathcal{D} : \{(x_i, y_i) | i = 1 \ldots N\} \sim P(x_i, y_i) \quad (1)$$

where $x_i \in R^{w \times h}$ is the $i$-th image, $y_i \in \{0, 1\}^L$ is the $i$-th observed noisy label vector, and $N$ is the number of samples. The noisy label $y_i$ is corrupted from the true label $y_i^* \in \{0, 1\}^L$ by an unknown process $y_i \sim P(y_i | x_i, y_i^*)$. In this work, we assume that we have a small portion of the dataset cleaned up, i.e., $\mathcal{D} = \mathcal{D}_c \cup \mathcal{D}_n$, where $\mathcal{D}_c$ is the small clean dataset and $\mathcal{D}_n$ is the remaining noisy data, and we have $N = |\mathcal{D}_c| + |\mathcal{D}_n|$ with $|\mathcal{D}_c| < |\mathcal{D}_n|$.

Our goal is to train an optimal classifier using the entire dataset $\mathcal{D}$. The classifier is optimal in the sense that the risk on unseen test data is minimized [21],

$$f^* = \mathrm{argmin}_f R_{\mathcal{D}_t}(f) = \mathrm{argmin}_f E_{\mathcal{D}_t}\{l[y^*, f(x)]\}, \quad (2)$$

where $f^*$ is the optimal classifier, $\mathcal{D}_t$ is the unseen test dataset, $y^*$ is the ground truth label of $x$, and $l[\cdot, \cdot]$ is a loss function, e.g., the commonly used cross entropy loss,

$$\begin{array}{l} l_{\mathrm{ce}}(y^*, f(x)) = -\sum_{m=1}^{L} \mathrm{CE}(y^*[m], \delta(f(x)[m])) \\ \mathrm{CE}(a, b) = a \ln b + (1-a) \ln(1-b), \end{array} \quad (3)$$

where $\delta(a) = 1/(1 + e^{-a})$ is the sigmoid activation, and $m$ in the bracket denotes the $m$-th element of the vector.

In the following sections, we introduce our distillation based algorithm to maximally leverage the partially labeled dataset $\mathcal{D}$, provide an analysis for the distillation algorithm, and extend the algorithm to use external knowledge graph information to further improve performance.

### 3.2. Knowledge Distillation

Our distillation framework is designed to be general, so that we do not rely on any particular assumption on label noise, because in practice the label noise is very diverse, non-stationary, and falls in multi-mode. Concretely, we first train an auxiliary model from the small clean dataset, and then transfer the knowledge learned from the auxiliary model to guide learning our primary model on the entire dataset. Our rationale is that the model trained from the small clean dataset produces an independent source of variance that can be used to cancel out the variance introduced by the label noise. We will have more analysis later in the section.

Given an auxiliary model $f_{\mathcal{D}_c}$ trained from the small clean dataset $\mathcal{D}_c$, we train our primary model with the entire dataset $\mathcal{D}$, using the following loss function [13, 9],

$$L_{\mathcal{D}}(y_i, f(x_i)) = \lambda l(y_i, f(x_i)) + (1-\lambda) l(s_i, f(x_i)), \quad (4)$$

where $s_i = \delta[f_{\mathcal{D}_c}(x_i)/T]$ and $T$ is the temperature [9]. In our experiments, we tried different temperatures, but the performance is not sensitive so we simply set $T = 1$ (See the sensitivity analysis in Table 5).

In Eqn. (4), the first term is the primary loss, and the second term is called the imitation loss [13]. The model is learned from the noisy labels with the primary loss, and at the same time to imitate the auxiliary model output $s_i$. $\lambda$ is a parameter to balance the noisy labels and the auxiliary model output. In the case of the cross entropy loss defined in Eqn. (3), the loss function is linear with respect to the label $y_i$, and Eqn. (4) can be rewritten as,

$$L_D(y_i, f(x_i)) = l(\lambda y_i + (1-\lambda) s_i, f(x_i)). \quad (5)$$

We define $\hat{y}_i^\lambda = \lambda y_i + (1-\lambda) s_i$ as the pseudo label, which combines the noisy label $y_i$ with the prediction of the auxiliary model output $s_i$. Both terms are deviated from the unknown true label $y_i^*$, but because the deviations are independent, this combined soft label can be closer to the true label under some conditions. By driving the pseudo labels closer to the ground truth label statistically, we can train a better model. We provide some analysis in the following.

**Rationale behind our distillation:** First, we define a risk $R_y$ associated with the unreliable label $\tilde{y}$:

$$R_{\tilde{y}} = E_{\mathcal{D}_t}[\|\tilde{y} - y^*\|^2], \quad (6)$$

where $y^*$ is the unknown ground truth label, and expectation is defined on the test set. The random variable $\tilde{y}$ denotes the unreliable label corrupted from the true label $y^*$, e.g., $s$ and $y$. Although $R_{\tilde{y}}$ does not relate directly with the final accuracy of the classifier, and the $\ell_2$ distance is different from the cross entropy loss function (3) [3], it is an indicator of the level of noise seen by the training process, which implicitly affects the final performance.

Next, we show that the risk of using the proposed pseudo label in Eqn. (5) can be smaller than using either the full noisy labels or only the partial clean labels. Specifically, we have the following proposition:

**Proposition 1.** *The optimal risk associated with $\hat{y}^\lambda$ is smaller than both risks with $y$ and $s$,* i.e.

$$\min_\lambda R_{\hat{y}^\lambda} < \min\{R_y, R_s\}, \quad (7)$$

---
[3] Ideally, we would like to define the risk according to the training loss, but $\ell_2$ distance is used for the tractability of analysis.

where $y$ is the unreliable label on $\mathcal{D}$, and $s$ is the soft label output from $f_{D_c}$. By setting $\lambda = \frac{R_s}{R_s+R_y}$, $R_{\hat{y}^\lambda}$ reaches its minimum,

$$\min_\lambda R_{\hat{y}^\lambda} = \frac{R_y R_s}{R_s + R_y}. \quad (8)$$

See proof in Appendix A. Eqn. (7) and (8) indicate that, by properly setting the balance weight $\lambda$, we can obtain a pseudo label $\hat{y}^\lambda$ that is closer to the ground truth label in the sense of $\ell_2$ distance. Therefore, we can potentially train a better classification model based on our distillation framework proposed in Eqn. (4).

Based on similar analysis as Eqn. (7), we can examine the effectiveness of other approaches including label smoothing [19] and bootstrap [16]. The label smoothing algorithm [19] revises the target label as,

$$\hat{y}^\lambda_u = \lambda y + (1-\lambda) u, \quad (9)$$

where $u$ is a vector of constants with each element set to $1/L$. Effectively, the label smoothing algorithm revises the noisy labels by damping the original $y$ and adding a uniform prior over all labels. Because $u$ is a constant, the independence assumption also holds, and the risk of the revised label can be reduced to,

$$\min_\lambda R_{\hat{y}^\lambda_u} = \frac{R_y R_u}{R_y + R_u}, \quad (10)$$

where $R_u = E_{\mathcal{D}_t}[\|u - y^*\|^2]$ is the risk of using a uniform distribution. It can be shown that the optimal risk in Eqn. (8) is smaller than that of Eqn. (10), i.e., $\min_\lambda R_{\hat{y}^\lambda} < \min_{\lambda_u} R_{\hat{y}^\lambda_u}$, if $R_s < R_u$. This is typically true because the auxiliary model is better than a uniform guess. Therefore, our distillation framework is more effective than the label smoothing algorithm, verified by our experiments as well.

The bootstrapping algorithm [16] revises the label as

$$\hat{y}^\lambda_{s'} = \lambda y + (1-\lambda) s', \quad (11)$$

where $y$ is the noisy label and $s'$ is the prediction from the current model in previous iteration. Because there is no additional side information used to train their model, $s'$ and $y$ are highly correlated, meaning that the revised label by bootstrap will have very similar risk as using the noisy label itself, and therefore, it is not as effective in handling labeling noise, which is also verified by our experiments.

### 3.3. Distillation Guided by Knowledge Graph

As the auxiliary model $f_{\mathcal{D}_c}$ is trained on a small clean dataset, it is highly likely to overfit to the small set of samples. To avoid over-certainty of the auxiliary model on its predictions [19], we propagate the label confidence among related labels to reduce the model variance for distillation. At the same time, different labels convey another sources of independent variances that might be beneficial for canceling out the labeling noise.

We leverage a knowledge graph $\mathcal{G}$ that encodes the structure of the label space. The knowledge graph has the form of a constrained matrix $G \in R_+^{L \times L}$, where $G(i,j)$ denotes the relationship between label $i$ and $j$, and $G(i,j) = 0$ indicates that the two labels are independent. We normalize the matrix such that each row sums up to one. We show how we use Wikipedia to construct such matrix as a knowledge graph in Section 4.1.

With this knowledge graph, we define the new soft label:

$$\hat{s}_i \triangleq G s_i, \quad (12)$$

based on the outputs of our auxiliary model. We then use the following loss function to train our primary model:

$$L_D(y_i, f(x_i)) = \lambda l(y_i, f(x_i)) + (1-\lambda) l(\hat{s}_i, f(x_i)), \quad (13)$$

where the soft label in Eqn. (4) is replaced with the new soft label $\hat{s}_i$ guided by our knowledge graph.

## 4. Datasets and Evaluation

In this section, we explain how we construct the noisy datasets in real-world scenarios and extensively compare our proposed approach with baselines on these datasets.

### 4.1. Datasets

Most existing work use simulated approaches to evaluate their method on learning from label noise [12, 18], where they inject label noise based on some controlled and known corruption process to the clean dataset. In contrast, our datasets reflect the practical setup: 1) Our datasets contain real-world label noise harvested from photo sharing sites. 2) Our datasets cover three domains of visual concepts, with varying levels of noise that come from different sources (e.g., text ambiguity such as polysemy words, real-world user behavior on photo tagging, etc.). 3) Background images are included in the evaluation set. Background images refer to images that do not belong to any of the classes in consideration. As shown in Krause *et al.* [10], if the evaluation set contains only the clean labeled images, the label noise does not affect the performance much for fine-grained classification. However, this is unlikely to be true in practice for image annotation – to evaluate our method in the real-world setting, we include background images into the evaluation.

We collect our training sets using Yahoo Flickr Creative Commons 100 Million (YFCC100M) [20], which is the largest public multimedia collection with a total of 100 million images and videos. YFCC100M provides a rich resource over a large amount of visual concepts, and reflects well the user preferences on the Flickr platform. However, compared with well adopted datasets, such as ImageNet, YFCC100M is missing clean annotations. Due to YFCC100M's large scale and real-world noisy labels, we believe it is a good resource to test our method. Figure 2 illustrates the overall workflow of collecting datasets from YFCC100M. Next, we explain how we build the noisy dataset to get $\mathcal{D}$ and $\mathcal{D}_c$.

| Reference | Teacher Network | Student Network |
|---|---|---|
| Hinton *et al.* [9] | Ensemble of strong *ConvNets* | Single fast *ConvNet* |
| Lopez-Paz *et al.* [13] | *ConvNet* with privileged features | *ConvNet* with generic features |
| Ours | *ConvNet* trained with clean dataset | *ConvNet* trained with noisy labels |

Table 1: Compare different distillation schemes.

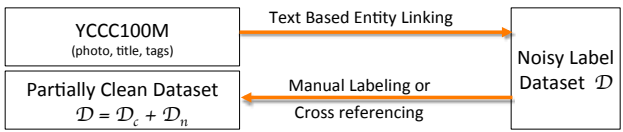

Figure 2: The data collection pipeline

| Domain | #Entity | #Photo |
|---|---|---|
| Place | 41,512 | 46,621,528 |
| People | 27,658 | 16,825,688 |
| **Species** | 5,958 | 4,086,366 |
| Work | 6,973 | 2,813,881 |
| **Artifacts** | 337 | 2,683,104 |
| **Sports** | 966 | 1,023,651 |
| Event | 1,519 | 1,088,173 |
| Food | 946 | 731,749 |
| Award | 57 | 32,724 |

Table 2: Statistics of the entity linking results on YFCC100M. Highlighted are those used in our evaluation.

From the YFCC100M dataset, we employ text based entity linking to connect images with the corresponding tags. We choose *DBpedia Spotlight* [5], an off-the-shelf tool that links a photo's title and tags with a Wikipedia entity. Given an ambiguous text term (*e.g.* apple), *DBpedia Spotlight* disambiguates different entities (*e.g.* Apple Inc. or apple the fruit) based on textual context. *DBpedia Spotlight* takes many text-related factors into consideration, such as synonyms and text morphological transformations. Based on the entity linking results, we label the photos into Wikipedia entities. This automatic labeling process produces the dataset $\mathcal{D}$ with label noise, where the label corruption process is unknown and complicated. Table 2 shows the statistics of the entity linking results for each domain.

We select three domains from the YFCC100M dataset, namely *Species*, *Sports* and *Artifacts*, which have enough training images and contain mostly visual entities. We avoid including certain domains where domain-specific model can excel, e.g., geo-tag based techniques can handle the *Places* domain. We also choose our domains by considering the overlap with ImageNet for cross-referencing.

The text-based entity linking introduces label noises for the dataset. To collect a partially clean dataset $\mathcal{D}_c$ on each domain, we try both crowdsourcing and automatic data linking approaches. For the *Species* and the *Sports* domains, we ask for the help of crowd-sourcing labeling from CrowdFlower [4] to clean part of the noisy labels. For *Species* and *Artifacts*, we cross-link their entities with ImageNet [17] synsets using the BabelNet dataset [15], which is a multilingual database linking various linguistic datasets, including WordNet and Wikipedia. We then use part of the images from ImageNet as the corresponding partial clean data. To differentiate differ-ent sources of clean data for the *Species* domain, we denote *Species-Y* and *Species-I* as two separate datasets where the partial clean data is from YFCC100M and ImageNet, respectively. Table 3 summarizes descriptive statistics of our datasets. We split each dataset into train/dev/test splits using the ratio of 6:3:1. We use the dev to select hyperparameters.

Figure 3 shows some examples images and their labels from our datasets. These examples show that the real-world label noises are caused by various reasons: 1) **Weak association**. Figure 3c is mistakenly labeled as the sport *Abseiling*, but the photo is not visually about *Abseiling* but shows a group of people, who are probably watching the sport event. 2) **Text ambiguity**. Figure 3p is mistakenly labeled as the species *Tulip*, but the image shows it was the texture pattern.

### 4.2. Implementation Details

Since we focus on the methodology of learning from noisy labels rather than squeezing the performance numbers, we use a simple variant of AlextNet [11] with batch normalization as the network for our evaluation. Adam optimizer is used to train the network. We train the network with 250 epochs, and with every 5 epochs, we reduce the learning rate by 0.9, and the initial learning rate is set to 0.001. During training, the performance is monitored based on the dev set to avoid overfitting. The datasets we collect are essentially multi-tag data. Instead of using top-K accuracy for evaluation, we use mean Average Precision (mAP) for the evaluation measurement.

As shown in the analysis of Section 3.2, the distillation parameters $\lambda$ and knowledge graph $G$ need to be properly chosen and designed, in order for the soft labels to achieve lower risk than the noisy labels. According to Eqn. (8), the optimal $\lambda$ can be computed according to the relative performance of the auxiliary model and the level of noise in the noisy dataset. Based on this principle, we use the following heuristics to find $\lambda$.

---
[4] http://www.crowdflower.com/

| Name | Clean Set $\mathcal{D}_c$ | Noisy Set $\mathcal{D}_n$ | $|\mathcal{D}_c|:|\mathcal{D}_n|$ | #Categories | #Train | #Dev | #Test |
|---|---|---|---|---|---|---|---|
| Sports | YFCC100M | YFCC100M | 1:1 | 238 | 86K | 18K | 52K |
| Species-Y | YFCC100M | YFCC100M | 1:1 | 219 | 50K | 10K | 28K |
| Species-I | ImageNet | YFCC100M | 1:4 | 219 | 93K | 14K | 40K |
| Artifacts | ImageNet | YFCC100M | 1:4 | 323 | 112K | 16K | 48K |

Table 3: Datasets statistics. The suffix (-Y and -I) refers to the source of the clean dataset $\mathcal{D}_c$.

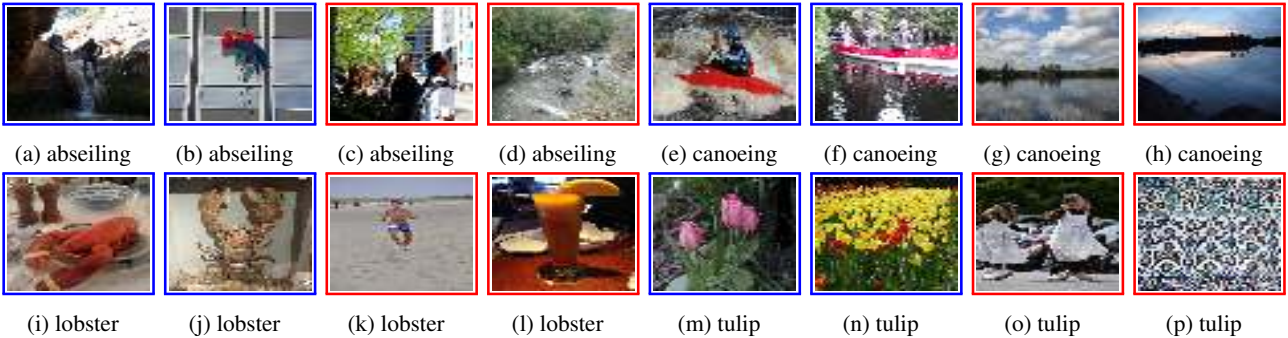

(a) abseiling (b) abseiling (c) abseiling (d) abseiling (e) canoeing (f) canoeing (g) canoeing (h) canoeing

(i) lobster (j) lobster (k) lobster (l) lobster (m) tulip (n) tulip (o) tulip (p) tulip

Figure 3: Example images and their noisy labels from our dataset. The three rows are from *Sports* and *Species*, respectively. The images with blue box are correctly labeled, and images with red box are mislabeled. The noisy labels are obtained by text based entity linking on the title and tag of the YFCC100M images. The noisily labeled images demonstrate various types of label noise seen in our dataset.

$$\lambda = \frac{\text{mAP}_{\mathcal{D}_c}}{\text{mAP}_{\mathcal{D}} + \text{mAP}_{\mathcal{D}_c}}, \quad (14)$$

where *mAP* score is computed from the dev set, and the subscript denotes the training dataset. The *mAP* is calculated as following:

$$mAP = \frac{1}{L}\sum_{i=1}^{L} AP_i, \quad (15)$$

where $AP_i$ is the average precision score for class $i$.

We employ predefined label relations to specify $G$, and one of the predefined label relations can be found easily at large scale on Wikipedia. We define the knowledge graph as $\mathcal{G} : (V, E)$, where $V$ denotes the entities, and the triplet $(u \in V, v \in V, r \in R) \in E$ denotes the entity relationship. $R$ denotes the type of relations in the knowledge graph. For example, on the *Species* domain, the top label relations are "class", "division", "family", "kingdom", "order" and "phylum", which are aligned with the tree of the life structure. On the *Sports* domain, the top label relations are "category", "equipment" and "genre". On the *Artifacts* domain, the top label relations are "type", "origin", "manufacturer", "genre","category", and "instrument". The number of relationship instances are 273, 2833 and 557 on *Sports*, *Species* and *Artifacts*, respectively. The directed edge $(u, v, r)$ means the entity $u$ is the relation $r$ of $v$. For example, ("Mammal", "Rabbit", "class") means "Mammal" is the "class" of "Rabbit".

Given the directed knowledge graph $\mathcal{G}$, the label relation matrix $G$ in the Eqn. (12) is defined as,

$$G(m, n) \propto \begin{cases} 1, m = n \\ \frac{\beta}{|\mathcal{N}(n)|}, if m \in \mathcal{N}(n) \\ 0, if (m, n) \notin E, \end{cases} \quad (16)$$

where $\mathcal{N}(n)$ denotes the *siblings* of the entities $n$ in the directed knowledge graph $\mathcal{G}$ and $\beta$ is a constant we set as 0.4 across different datasets.

### 4.3. Qualitative Results

The key of the proposed distillation framework is to drive the pseudo label $\hat{y}_i = \lambda y_i + (1-\lambda)s_i$ and the guided pseudo label by knowledge graph $\hat{y}_i^g = \lambda y_i + (1-\lambda)\hat{s}_i$ statistically closer to the unknown true label $y_i^*$. Figure 4 show examples of the class *Bison* from the *Species* domain, assuming it is the $m^{th}$ class. $y_i[m]$, the observed labels, is all 1 for these images, *i.e.* only the observed positive images are shown in the example. $y_i^*[m]$, the hidden ground truth labels, is illustrated by the color of the boxes, the red box means the $y_i^*[m] = 0$, and the blue box means the $y_i^*[m] = 1$. The first row is ranked by elements in $\hat{y}_i$ of Eqn. (5), the distillation pseudo labels. The second row is ranked by elements in $\hat{y}_i^g$ of Eqn. (12), the guided distillation pseudo labels.

|  | Sports | Species-Y | Species-I | Artifacts |
|---|---|---|---|---|
| Baseline-Clean | 44.0 | 18.1 | 22.0 | 19.2 |
| Baseline-Noisy [10] | 50.7 | 23.7 | 38.5 | 22.0 |
| Baseline-Ensemble | 52.2 | 25.1 | 39.1 | **26.9** |
| Bootstrap [16] | 50.6 | 23.6 | 38.8 | 23.4 |
| Label Smooth [19] | 51.9 | 25.1 | 41.4 | 22.9 |
| Finetune | 50.8 | 22.2 | 37.5 | 19.7 |
| Noise Layer [18] | 50.8 | 23.7 | 38.5 | 22.0 |
| Importance Re-weighting [12] | 50.8 | 23.7 | 41.6 | 24.8 |
| Distillation (Eqn. (4)) | 53.5 | **26.1** | 41.6 | 26.0 |
| Semantic Guided Distillation (Eqn. (13)) | **53.7** | 25.2 | **42.3** | 26.0 |
| Upper Bound | 54.1 | 27.4 | - | - |

Table 4: Experiment results. The numbers are the mAP scores (%) defined in Eqn. (15).

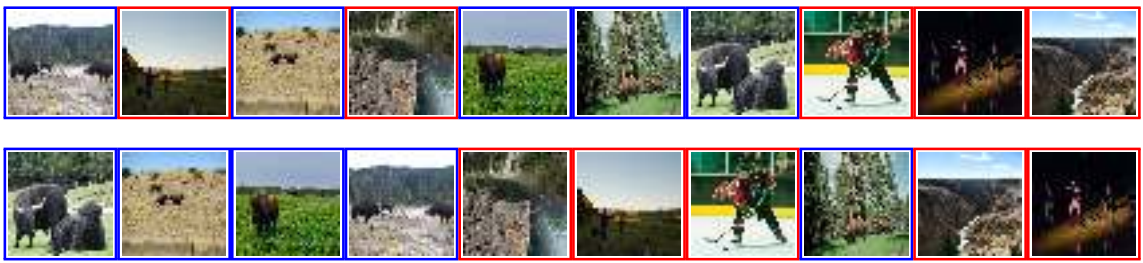

Figure 4: Ranking noisy dataset $\mathcal{D}$ with pseudo labels. All images are observed as the class *Bison* of the *Species* domain. The two rows, with the same set of images, are ranked by the distillation and guided distillation pseudo labels, respectively. The red and blue borders denotes noise/clean observations, respectively. Analysis in Section 4.3.

Figure 4 shows that the $\hat{y}_i$ improves over $y_i$ by ranking the true positives higher, and false positives lower, and the guided pseudo labels $\hat{y}_i^g$ further improve over $y_i$ by picking up confidences from related classes, and ranked more true positives higher.

### 4.4. Comparison

Table 4 shows the experimental results. We can see that our distillation method and graph guided distillation consistently outperforms all baselines and existing methods. Next we will discuss the comparisons in details.

**Comparing with Baselines.** Following Krause *et al.* [10], we define *Baseline-Clean* and *Baseline-Noisy* to denote the baseline models trained with the partial clean dataset $\mathcal{D}_c$ and the entire noisy dataset $\mathcal{D}$, respectively. These models share the same network structure as our distillation model, except that the input of labels during training is different. The model trained with *Baseline-Clean* is used as the auxiliary model (e.g., teacher model) in our distillation method. Our new model consistently outperforms these two models. Especially, if we view our new model as the student model and the *Baseline-Clean* as the teacher model, the student model is better than the teacher. This is because the student learns from more training examples, and it leverage the teacher model to reduce the effect of noisy label. We also compare with the *Baseline Ensemble*, which combines the two baseline models *Baseline-Clean* and *Baseline-Noise* by geometric mean. The results show that our distillation model is better than *Baseline Ensemble* in three of the four datasets, and works comparably on the last one. However, our method has a big advantage over the *Baseline Ensemble* in that we only need one forward pass on our CNN model while *Baseline Ensemble* need to run two CNN models, which is twice time consuming as well as doubles storage compared to ours.

**Comparing with Bootstrapping and Label Smoothing.** The *Bootstrapping* method [16] and *Label Smoothing* [19] are popular approaches to reduce the noise effects. Section 3.2 has discussed in theory the advantage of our distillation based approach against these two approaches. The experiment results verify that our method consistently outperforms these two approaches in all the four datasets.

**Comparing with Finetuning, Importance Re-weighting, and Noisy Layer.** As discussed in the related work section, there are a number of practices in the recent deep learning literature improving the performance with noisy labels. The simplest one is *Finetune*, which initialize network weights from the *Baseline-Clean* model, and subsequently finetune using the entire noisy dataset $\mathcal{D}_n$. To further reduce the noise label effects, *Importance Re-weighting* [12] introduces estimated weights on the noisy labels, while *Noisy Layer* [18] employs an extra linear layer on top of the network to "absorb" label switch noise. Among these three approaches, *Importance Re-weighting* outperforms the other two, but is still inferior to our method with a significant gap.

| $T$ | 0.5 | 1 | 5 | 10 |
|---|---|---|---|---|
| mAP(%) | 41.5 | 41.6 | 39.8 | 35.3 |

Table 5: mAP performance using different $T$ on the *Species-I* dataset

**Comparing Distillation and Guided Distillation with Upper Bound.** Inevitably, the label noise will hurt performance, compared with using full clean dataset. On the *Species-Y* and *Sports* datasets, where the clean datasets are collected by crowd sourcing, we continue to label the rest of the noisy dataset $\mathcal{D}$, and use the fully labeled dataset to train a *ConvNet*. The performance of this *ConvNet* can be seen as an "Upper Bound" of the learning with noisy labels, because the goal of learning with noisy labels is to approach the performance of the model trained with the fully cleaned dataset. For the *Sports* and *Species-Y* datasets, the proposed guided distillation is very close to the upper bound, which means the proposed distillation method can save the budget of labeling the rest of the dataset.

**Comparing Different Temperature** $T$ We perform sensitivity analysis of the hyper-parameter, *i.e.*, the temperature $T$, in Eqn. (4) on the *Species-I* dataset. The results are listed in Table 5, which shows that the performances with different small temperatures $T$ is stable.

## 5. Conclusion

This paper developed a new framework to learn from noisy labels, by leveraging the knowledge learned from a small clean dataset and semantic knowledge graph to correct the noisy labels. To standardize the evaluation protocol for systems that learn from noisy labels, we collected a suite of new datasets in Species, Sports and Artifacts domains, which reflect the real-world labeling noise. The proposed methods not only produce superior performance on the task of learning from noisy data, but also provide unique and novel perspectives on the distillation framework.

Moving forward, we intend to explore in distillation with new source of guidance in addition to knowledge graphs. We are also interested in applying our method to other scenarios with noisy labels, such as Web-scale photo search.

## A. Proof for Proposition 1

*Proof.* The risk of using labels corrupted by noise as $y \sim P_\mathcal{D}(y|(x, y^*))$ is quantified by the following residual term,

$$R_y = E_{\mathcal{D}_t}[\|y - y^*\|^2], \quad (17)$$

Consider our auxiliary model $f_{\mathcal{D}_c}$ trained from a clean dataset $\mathcal{D}_c$, the expected prediction error can be decomposed into the variance term and the bias term [7],

$$E_{\mathcal{D}_t}[l(s, y^*)] = l(\bar{s}, y^*) + E_{\mathcal{D}_t}[l(s, \bar{s})], \quad (18)$$

where $l(\cdot, \cdot)$ is the loss function, and $\bar{s}(x)$ is called the "main" prediction, defined according to the loss function. For the squared loss $\bar{s}(x)^{\text{sq}} = \text{average}_{\mathcal{D}_t}(f_{\mathcal{D}_c}(x))$, and for the 0-1 loss $\bar{s}(x)^{\text{0-1}} = \text{median}_{\mathcal{D}_t}(f_{\mathcal{D}_c}(x))$ [7]. For simplicity of proving Eqn. (7), we use the squared loss. Since we are training a high capacity CNN model, we can make a reasonable assumption that the bias term $l(\bar{s}, y^*)$ is close to zero. Therefore we have,

$$\begin{aligned} l(\bar{s}, y^*) \approx 0 \Rightarrow \bar{s} \approx y^* \\ E_{\mathcal{D}_t}(\|s - y^*\|^2) \approx E_{\mathcal{D}_t}(\|s - \bar{s}\|^2) \triangleq R_s. \end{aligned} \quad (19)$$

The label corruption process is unknown, but we can assume that it is independent of the model variance. This leads to

$$\begin{aligned} E_{\mathcal{D}_t}[(y - y^*)^T(s - y^*)] &= [E_{\mathcal{D}_t}[y - y^*]]^T E_{\mathcal{D}_t}[s - y^*] \\ \because \text{Eqn. (19) } \bar{s} \approx y^* \Rightarrow &= [E_{\mathcal{D}_t}[y - y^*]]^T E_{\mathcal{D}_t}[s - \bar{s}] \\ \because E_{\mathcal{D}_t}(s) = \bar{s} \Rightarrow &= [E_{\mathcal{D}_t}(y - y^*)]^T \mathbf{0} = 0, \end{aligned} \quad (20)$$

where $\mathbf{0}$ denotes a zero vector.

Now, we are ready to show Eqn. (7),

$$\begin{aligned} R_{\hat{y}^\lambda} &= E_{\mathcal{D}_t}[\|\hat{y}^\lambda - y^*\|^2] \\ &= E_{\mathcal{D}_t}[\|\lambda y + (1-\lambda)s - y^*\|^2] \\ &= E_{\mathcal{D}_t}[\|\lambda(y - y^*) + (1-\lambda)(s - y^*)\|^2] \\ \because \text{Eqn. (20)} &= \lambda^2 R_y + (1-\lambda)^2 R_s, \end{aligned} \quad (21)$$

By setting $\lambda = \frac{R_s}{R_s + R_y}$, $R_{\hat{y}^\lambda}$ reaches its minimum,

$$\min_\lambda R_{\hat{y}^\lambda} = \frac{R_y R_s}{R_s + R_y}, \quad (22)$$

which concludes the proof of Proposition 1. □